\documentclass[sigconf]{acmart}
\usepackage{amsmath,amsfonts}
\usepackage{graphicx}
\usepackage{xcolor}
\usepackage[export]{adjustbox}
\usepackage{amstext}
\usepackage{cuted}
\usepackage{capt-of}
\usepackage{multirow}
\usepackage{mathtools} 

\AtBeginDocument{%
  }

\copyrightyear{2026}
\acmYear{2026}
\setcopyright{cc}
\setcctype{by}
\acmConference[MM '26]{Proceedings of the 34th ACM International Conference on Multimedia}{November 10--14, 2026}{Rio de Janeiro, Brazil}

\acmBooktitle{Proceedings of the 34th ACM International Conference on Multimedia (MM '26), November 10--14, 2026, Rio de Janeiro, Brazil}
\acmDOI{10.1145/3767308.3832539}
\acmISBN{979-8-4007-2213-4/2026/11}

\begin{document}

\title{Towards Adaptive Super-Resolution and Quality Assessment via Test-Time Adaptation}

\author{Ajeet Kumar Verma}
\email{ajeet.verma@iitjammu.ac.in}
\orcid{0000-0003-1419-0557}
\affiliation{%
  \institution{Indian Institute of Technology Jammu}
  \city{Jammu}
  \state{J \& K}
  \country{India}
}

\renewcommand{\shortauthors}{Ajeet Kumar Verma}

\begin{strip}
  \centering
  \begin{minipage}{\textwidth}
    \centering
    \includegraphics[width=\textwidth]{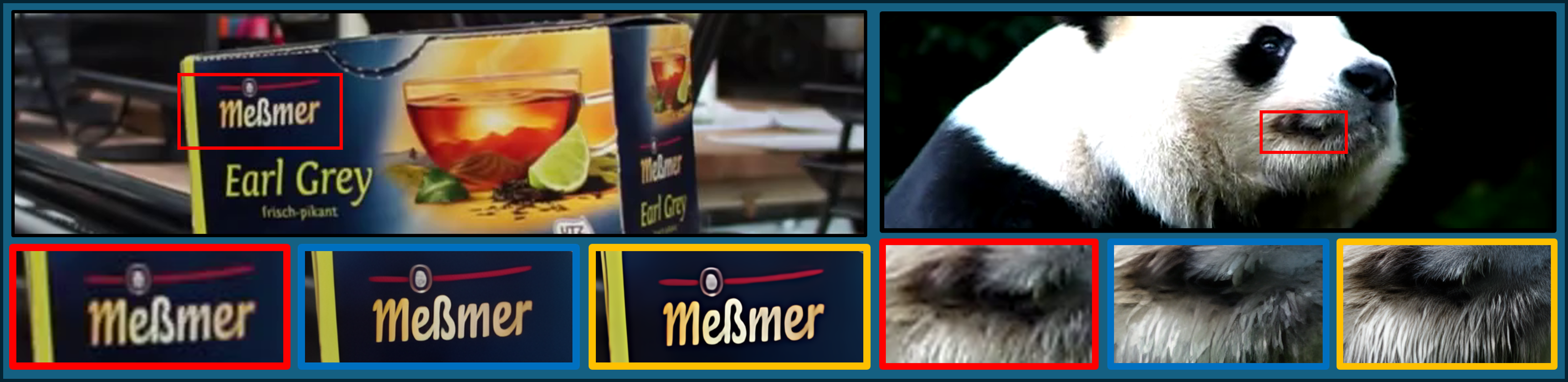}
  \end{minipage}
  \captionof{figure}
  {Qualitative comparison between Baseline and its adapted version. The cropped LR region is highlighted in \textcolor{red}{red}, the base model output in \textcolor{blue}{blue}, and the adapted model output with adaptive quality loss in \textcolor{yellow}{yellow} [Zoom-in for best view].} 
  \label{fig:teaser}
\end{strip}

\begin{abstract}
This paper presents doctoral research on adaptive video super-resolution and perceptual quality modeling under real-world conditions. Existing video super-resolution (VSR) methods struggle to generalize under unknown degradations arising from heterogeneous devices, codecs, and network environments.
We address this challenge through test-time adaptation (TTA), a unified paradigm that improves robustness and perceptual quality without retraining or high-quality supervision. Specifically, we: 1) propose a TTA-based framework for no-reference video quality assessment (VQA), where adapted quality predictions provide perceptual guidance for VSR under unseen distortions; 2) develop a transformer-based architecture for screen-content super-resolution that preserves text clarity and structural fidelity; and 3) introduce a region-aware TTA strategy that selectively refines text and non-text regions without requiring high-resolution ground truth.
Experimental results across diverse benchmarks demonstrate consistent improvements in perceptual quality and readability. We also outline ongoing work toward fully adaptive video enhancement systems capable of generalizing across unseen domains.
\end{abstract}



\maketitle

\section{Introduction}

Real-world videos are very common in modern communication systems, including streaming platforms, remote collaboration, and online learning. In practice, videos often suffer from complex, unknown degradations, such as compression artifacts, noise, motion blur, temporal inconsistencies, and scaling distortions, that vary across devices, codecs, and network conditions, posing challenges for learning-based models trained under controlled assumptions.

VSR attempts to reconstruct high-resolution (HR) videos from degraded low-resolution (LR) inputs. Deep learning approaches\cite{edsr, tecoGAN, basicvsrpp} perform well on benchmarks but are trained on paired LR–HR data with fixed degradation models and losses. Consequently, their performance drops on real-world videos with mixed, unseen distortions. Evaluating perceptual quality is also difficult: full-reference metrics\cite{dists, afine, eiqm, dfss} require HR data, while no-reference VQA methods\cite{2022faster_vqa, fast_vqa, dover, cover, sama, arniqa} often fail to generalize to super-resolved videos under complex distortions.

Screen content adds further challenges, as text, sharp edges, and structured graphics are highly sensitive to distortions. This doctoral research proposes adaptive, reference-free frameworks that: (i) improve real-world VSR with perceptually guided optimization, (ii) enhance screen content VSR with content-aware modeling, and (iii) enable TTA for screen content SR, focusing on text and structural fidelity for robust, perceptually aligned results.

\section{Motivation and Problem Statement}

Despite advances in super-resolution and video quality assessment, existing methods struggle in real-world scenarios. Most SR models rely on paired LR–HR data with fixed degradations, which fail to capture practical distortions. Applied to compressed videos or screen content, they often produce over-smoothed textures, amplified artifacts, or distorted edges. Additionally, the lack of high-resolution reference data prevents post-training adaptation.

Current quality assessment (QA) methods\cite{2022faster_vqa, fast_vqa, dover, cover, sama, arniqa} offer limited guidance: full-reference metrics are impractical, and no-reference models often misalign with human perception under mixed distortions. As a result, SR pipelines lack reliable perceptual feedback for adaptive optimization beyond pixel-level fidelity.

These limitations reveal two critical gaps: (i) the need for SR frameworks that integrate adaptive, no-reference perceptual guidance to handle unknown degradations in real-world videos, and (ii) the need for specialized strategies for screen content, where preserving text clarity, edge sharpness, and structural fidelity is essential.

To address these gaps, we propose three key frameworks:  
\begin{itemize}
    \item \textit{Adaptive no-reference video quality assessment for RW-VSR}: TTA of VQA models provides perceptual feedback aligned with human judgment, which is then used as an adaptive loss to guide real-world video super-resolution\cite{rwvsrqa}.
    \item \textit{Screen content video super-resolution (ScrVSR)}: A content-aware SR model tailored for screen recordings, emphasizing text readability, edge preservation, and robustness to H.265 compression\cite{scrvsr}.
    \item \textit{TTA for screen content SR (SCISR-TTA)}: A dual-branch adaptation framework for screen content images that refines text and non-text regions separately during inference, improving perceptual quality without ground-truth references \cite{scisrtta}.
\end{itemize}

By explicitly addressing these gaps, we propose advancements to SR frameworks that are robust, perceptually guided, and practical for deployment across both natural and screen content.

\section{Methodology and Results}
In this section, we provide brief details on addressing the identified gaps and the associated results.

\subsection{Test-Time Adaptation for RW-VSR-QA Algorithms and an adaptive framework to guide RW-VSR Models}

\begin{figure}[ht]
    \centering
    \includegraphics[width=\linewidth]{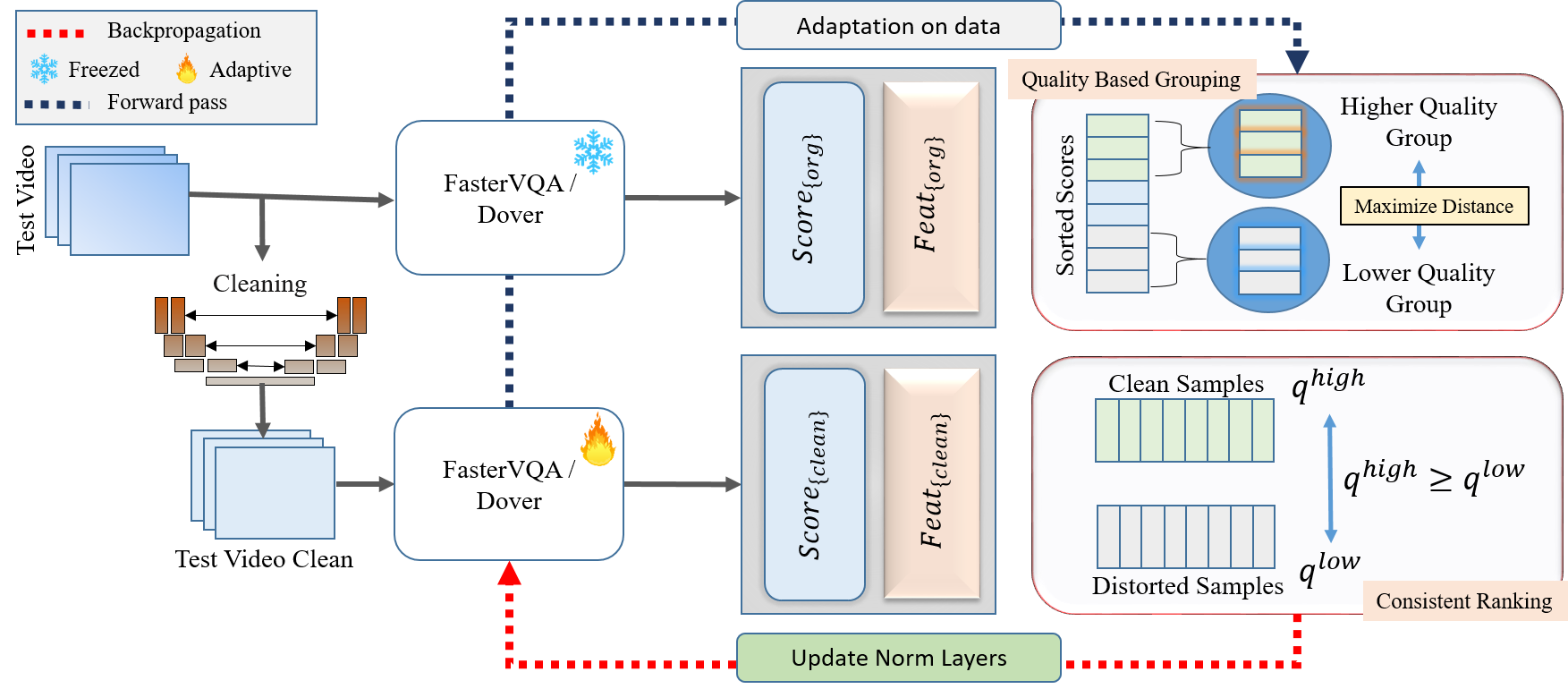}
    \caption{Overview of the proposed TTA-based RW-VSR-QA algorithm, in which TTA incorporates group contrastive loss and quality-aware rank loss
    as auxiliary tasks to improve performance of the video quality assessment algorithm under real-world degradations.}
    \label{fig:model_arc_tta}
\end{figure}

\textbf{Method Part-I}: In this work, we apply TTA to real-world video quality assessment to improve robustness under unknown and mixed degradations. In the proposed TTA framework (Shown in Figure \ref{fig:model_arc_tta}, each batch of test video clips is used to update a small subset of the QA network's parameters, thereby better aligning the model with the current test data distribution. Specifically, only the normalization layer parameters are updated. This design enables the model to respond to distribution shifts while preserving the semantic knowledge learned during training. To update the model, we use two auxiliary learning objectives, namely Quality-Based Group Contrastive Loss and Quality-Aware Rank Loss (QARL).

\noindent\textbf{Results}:
The proposed framework improves VQA in a fully unsupervised manner, without requiring ground-truth subjective labels or retraining of the original models. This is crucial for RW-VSR, where collecting human opinion scores for every new distortion is impractical. By adapting directly to test data, the pre-trained VQA model better aligns with the statistics and perceptual characteristics of super-resolved real-world videos.

\begin{table}[ht]
\centering
\caption{Performance comparison of base VQA methods and their adapted versions using the proposed TTA algorithm,  Symbol $\uparrow$ represents higher is better and $\downarrow$ represents lower is better. }
\resizebox{\columnwidth}{!}{%
\begin{tabular}{cc||c|c|c|c|}
\hline
\multicolumn{2}{|c||}{\textbf{VQA Baseline}} & \textbf{PLCC$\uparrow$} & \textbf{SRCC$\uparrow$} & \textbf{KRCC$\uparrow$} & \textbf{RMSE$\downarrow$} \\ \hline \hline
\multicolumn{1}{|c|}{\multirow{2}{*}{FVQA\cite{2022faster_vqa}}} & Baseline & 0.6021 & 0.5922 & 0.4100 & 15.9417 \\ \cline{2-6} 
\multicolumn{1}{|c|}{} & TTA-VQA & \textbf{0.6167} & \textbf{0.6038} & \textbf{0.4197} & \textbf{15.7167} \\ \hline 
\multicolumn{1}{|c|}{\multirow{2}{*}{Fast-B\cite{fast_vqa}}} & Baseline & 0.6826 & 0.6490 & 0.4451 & 14.5912 \\ \cline{2-6} 
\multicolumn{1}{|c|}{} & TTA-VQA & \textbf{0.6976} & \textbf{0.6609} & \textbf{0.4549} & \textbf{14.3060 }\\ \hline
\multicolumn{1}{|c|}{\multirow{2}{*}{Dover\cite{dover}}} & Baseline & 0.6274 & 0.5998 & 0.4033 & 15.5442 \\ \cline{2-6} 
\multicolumn{1}{|c|}{} & TTA-VQA &\textbf{ 0.6308} &\textbf{ 0.6113} & \textbf{0.4309} & \textbf{15.4921} \\ \hline  
\multicolumn{1}{|c|}{\multirow{2}{*}{COVER\cite{cover}}} & Baseline & 0.7501 & 0.6897 & 0.4718 & 13.2038 \\ \cline{2-6} 
\multicolumn{1}{|c|}{} & TTA-VQA &\textbf{ 0.7542} & \textbf{0.6928} & \textbf{0.4750} & \textbf{13.1094} \\ \hline 
\multicolumn{1}{|c|}{\multirow{2}{*}{SAMA\cite{sama}}} & Baseline &  0.7265 & 0.6930 &  0.4849 & 13.7201 \\ \cline{2-6} 
\multicolumn{1}{|c|}{} & TTA-VQA &\textbf{ 0.7767} & \textbf{0.7471} & \textbf{0.5357} &  \textbf{12.5757} \\ \hline 

\label{table:tta_vqa}
\end{tabular}
}
\end{table}

As observed in Table \ref{table:tta_vqa}, across all baselines, TTA consistently enhances performance. SRCC gains of 1.92\%, 1.80\%, 1.88\%, 0.45\%, and 7.24\% are observed for Faster-VQA\cite{2022faster_vqa}, FAST-VQA\cite{fast_vqa}, DOVER\cite{dover}, COVER\cite{cover}, and SAMA\cite{sama}, respectively. These improvements indicate a more reliable quality ranking, essential for tasks such as model selection and adaptive streaming. The largest gain for SAMA\cite{sama} highlights the benefit of TTA for lightweight single-branch models under complex real-world distortions.

\noindent\textbf{Method Part-II}: RW-VSR often struggles with unknown degradations such as motion blur, noise, compression artifacts, and mixed distortions. Conventional VSR training uses fixed, pixel-based, or perceptual losses that cannot capture the impact of these distortions, and designing separate losses for each type is impractical.
To address this, we introduce an adaptive loss based on the RW-VSR-QA test-time adaptation framework. The quality model is first adapted to super-resolved videos to learn their distortions and accurately predict perceptual quality. An efficient SR model should generate videos whose quality scores closely match those of HR videos; the difference between these scores forms the adaptive quality loss.

\begin{figure}[ht]
    \centering
    \includegraphics[width=\linewidth]{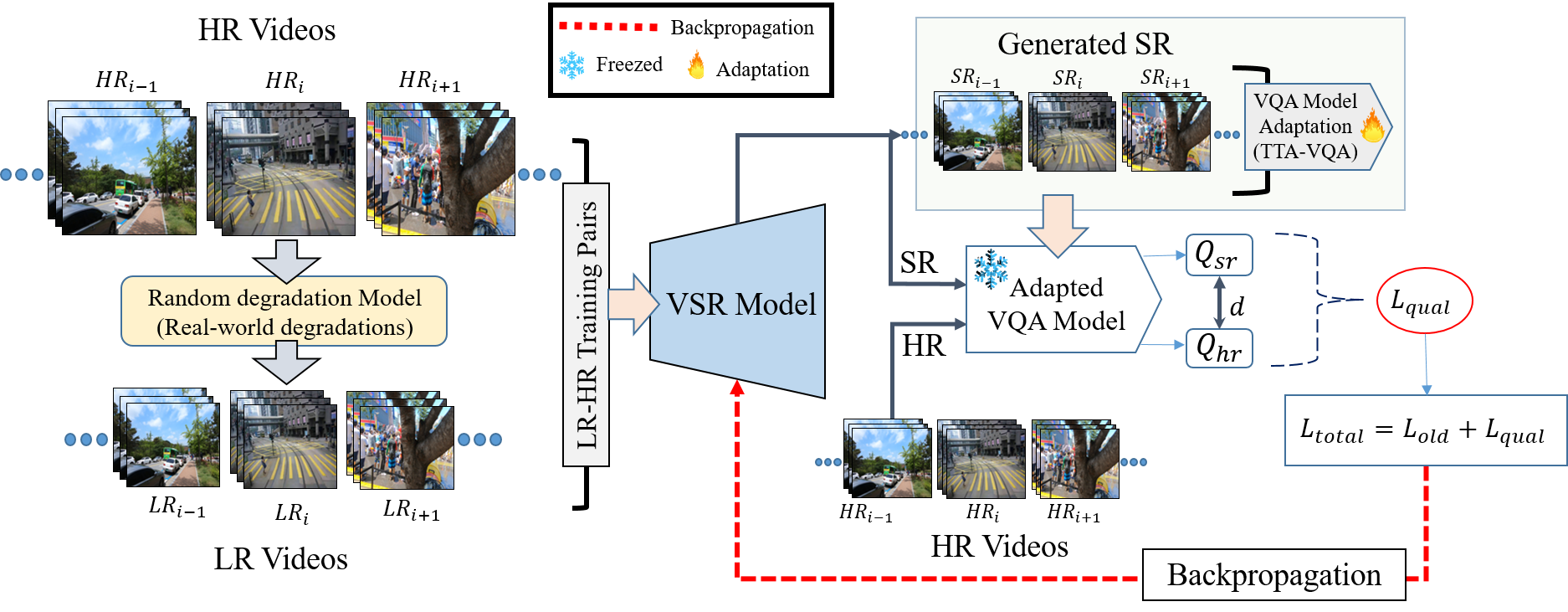}
    \caption{Overview of the proposed RW-VSR framework using the RW-VSR-QA algorithm as an adaptive loss function. Red dotted lines indicate the weightupdate process, and d represents the quality difference.}
    \Description{Diagram showing the RW-VSR framework and its components.}
    \label{fig:vsr_tta}
\end{figure}

\noindent \textbf{Results}:
The proposed framework is evaluated on six representative VSR backbones, namely BasicVSR\cite{basicvsrpp}, RBVSR\cite{real_basic_2022}, HAT\cite{hat}, SRWD\cite{srwd}, RESR\cite{realesrgan}, and IART\cite{iart}, across four challenging real-world datasets: RVSR, VLQ, K$|$Lens, and MotionBlur. 

\begin{table}[ht]
\centering
\caption{Quantitative comparison of various VSR models and their enhanced versions using the proposed RW-VSR-QA algorithm as an adaptive loss. Base VSR models are cited by name, with `\_Q'  indicating models integrated with adaptive quality loss. Higher values represent better results. (MB: MotionBlur)}

\begin{tabular}{|l||rrrr|r|}
\hline
\multirow{2}{*}{\textbf{VSR Model}} & \multicolumn{4}{c|}{\textbf{Dataset}} & \multicolumn{1}{l|}{\multirow{2}{*}{\textbf{Avg.}$\uparrow$}} \\ \cline{2-5}
 & \multicolumn{1}{c|}{\textbf{RVSR}} & \multicolumn{1}{c|}{\textbf{MB}} & \multicolumn{1}{c|}{\textbf{VLQ}} & \multicolumn{1}{c|}{\textbf{K$|$Lens}} & \multicolumn{1}{l|}{} \\ \hline \hline
BasicVSR\cite{basicvsrpp} & \multicolumn{1}{c|}{0.2006} & \multicolumn{1}{c|}{0.4945} & \multicolumn{1}{c|}{0.4853} & 0.3141 & 0.3736 \\ \hline
BasicVSR\_Q & \multicolumn{1}{c|}{\textbf{0.2040}} & \multicolumn{1}{c|}{\textbf{0.5018}} & \multicolumn{1}{c|}{\textbf{0.4931}} & \textbf{0.3174} & \textbf{0.3791} \\ \hline
RBVSR\cite{real_basic_2022} & \multicolumn{1}{c|}{0.7813} & \multicolumn{1}{c|}{0.6925} & \multicolumn{1}{c|}{0.7356} & 0.6419 & 0.7128 \\ \hline
RBVSR\_Q & \multicolumn{1}{c|}{\textbf{0.7825}} & \multicolumn{1}{c|}{\textbf{0.6928}} & \multicolumn{1}{c|}{\textbf{0.7596}} & \textbf{0.6711} & \textbf{0.7265} \\ \hline
HAT\cite{hat} & \multicolumn{1}{c|}{0.7765} & \multicolumn{1}{c|}{0.6392} & \multicolumn{1}{c|}{0.7474} & 0.5898 & 0.6883 \\ \hline
HAT\_Q & \multicolumn{1}{c|}{\textbf{0.7766}} & \multicolumn{1}{c|}{\textbf{0.6697}} & \multicolumn{1}{c|}{\textbf{0.7584}} & \textbf{0.6378} & \textbf{0.7106} \\ \hline
SRWD\cite{srwd} & \multicolumn{1}{c|}{0.7693} & \multicolumn{1}{c|}{0.6458} & \multicolumn{1}{c|}{\textbf{0.7549}} & 0.6516 & 0.7054 \\ \hline
SRWD\_Q & \multicolumn{1}{c|}{\textbf{0.7723}} & \multicolumn{1}{c|}{\textbf{0.7131}} & \multicolumn{1}{c|}{0.7523} & \textbf{0.6871} & \textbf{0.7312} \\ \hline
RESR\cite{realesrgan} & \multicolumn{1}{c|}{0.7539} & \multicolumn{1}{c|}{0.6571} & \multicolumn{1}{c|}{0.7023} & 0.6254 & 0.6847 \\ \hline
RESR\_Q & \multicolumn{1}{c|}{\textbf{0.7798}} & \multicolumn{1}{c|}{\textbf{0.6837}} & \multicolumn{1}{c|}{\textbf{0.7445}} & \textbf{0.6740} & \textbf{0.7205} \\ \hline
IART\cite{iart} & \multicolumn{1}{c|}{0.1932} & \multicolumn{1}{c|}{0.4955} & \multicolumn{1}{c|}{0.5595} & 0.3457 & 0.3985 \\ \hline
IART\_Q & \multicolumn{1}{c|}{\textbf{0.2327}} & \multicolumn{1}{c|}{\textbf{0.5028}} & \multicolumn{1}{c|}{\textbf{0.5641}} & \textbf{0.3757} & \textbf{0.4188} \\ \hline
\end{tabular}

\label{table:vsr_tta}
\end{table}

Results in Table \ref{table:vsr_tta} and Figure \ref{fig:teaser} show that integrating RW-VSR-QA as an adaptive quality loss consistently improves performance across all backbones and datasets, complementing conventional pixel and perceptual losses. Improvements are most pronounced on datasets with severe real-world distortions, such as MotionBlur and K$|$Lens. The gains are observed and rely heavily on perceptual and adversarial objectives that can amplify high-frequency noise. Incorporating RW-VSR-QA guides these models toward visually clean and stable outputs, resulting in perceptual quality improvements. This work has been published in IEEE Transactions on Artificial Intelligence\cite{rwvsrqa}.

\subsection{Super-Resolution for Screen Content Videos (ScrVSR)}
\textbf{Methodology}: 
The proposed model, \textbf{ScrVSR}, extends the baseline ITSRN\cite{itsrn} architecture with enhancements tailored for screen-sharing videos. It leverages pretrained ITSRN weights to accelerate convergence and utilize transformer-based continuous image representations. Unlike conventional super-resolution models that focus on aesthetic enhancements, ScrVSR explicitly preserves text clarity, edge sharpness, and structural fidelity, which are critical for screen content. To achieve this, we employ a text-aware, perceptual-quality loss framework. Specifically, a character error rate loss, which ensures that textual content is reconstructed accurately, while a CLIP-based quality loss provides perceptual feedback that aligns super-resolved frames with high-resolution ground truth. Additionally, a VGG-based perceptual loss encourages high-level structural and semantic consistency.
Training emphasizes text-rich regions, enabling the network to reconstruct high-fidelity text, crisp edges, and fine graphical details while maintaining robustness to compression artifacts and low-resolution inputs, making it well-suited for real-world screen content super-resolution.

\begin{figure}[ht]
    \centering
    \includegraphics[width=\linewidth]{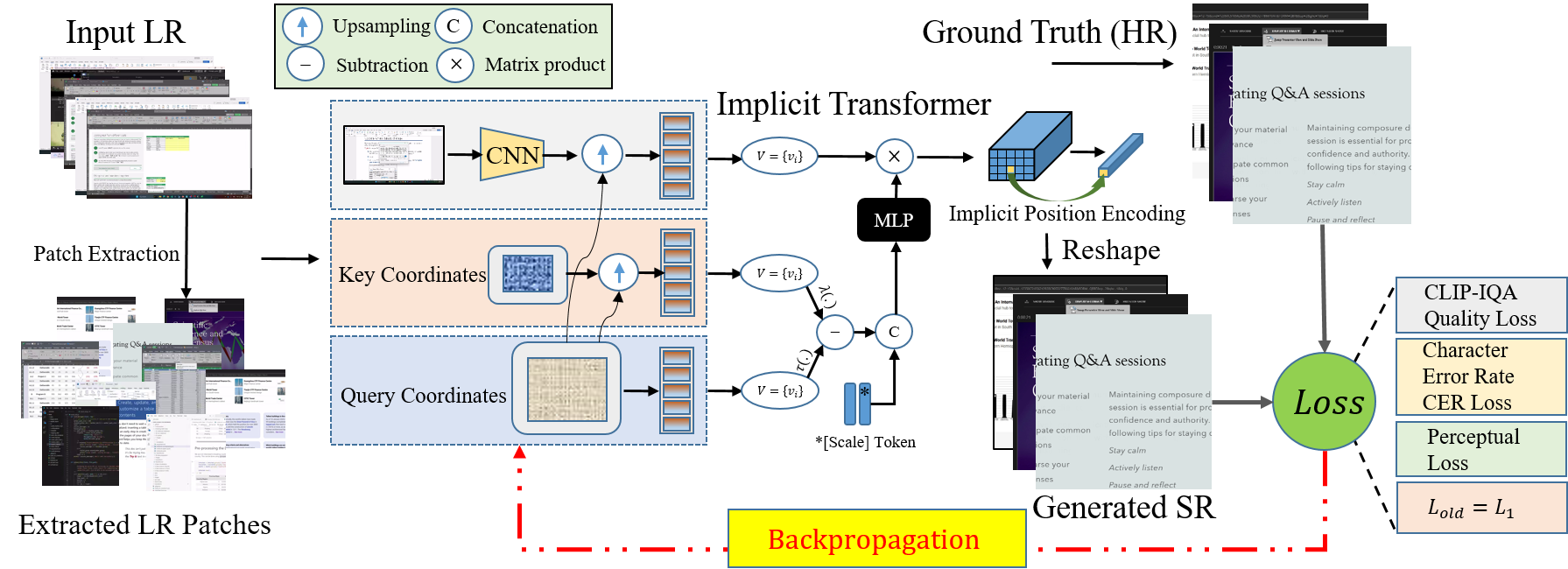}
    \caption{Overview of the proposed ScrVSR model. Training is supervised; after pretrained initialization, losses are computed between the high-resolution (HR) and super-resolved (SR) outputs. All losses are combined in a regularized manner and propagated backward.}
    \label{fig:ScrVSR}
\end{figure}

\textbf{Results}:
 The performance is assessed using three complementary metrics: Peak Signal-to-Noise Ratio (PSNR), Structural Similarity Index (SSIM), and Character Error Rate (CER). Higher Quantization Parameter (QP) values correspond to increased compression and, consequently, greater degradation in the input video quality, making restoration progressively more challenging.
The results demonstrate that ScrVSR consistently outperforms baseline methods across nearly all quantization settings.

\begin{table}[ht]
    \centering
    \caption{\textbf{Comparison of the proposed model with BTC and ITSRN across different quantization settings}. Higher values are better for PSNR and SSIM ($\uparrow$), and lower values are better for CER ($\downarrow$).}
    \label{tab:qp_wise_comparison}
    \begin{tabular}{|c||l|c|c|c|}
    \hline
    \textbf{Quant Setting} & \textbf{MODEL} & \textbf{PSNR$\uparrow$} & \textbf{SSIM$\uparrow$} & \textbf{CER$\downarrow$} \\ \hline \hline

    \multirow{3}{*}{QP-17} 
    & BTC\cite{btc} & 27.18 & 0.9449 & 0.2137 \\ \cline{2-5}
    & ITSRN\cite{itsrn} & 28.31 & 0.9548 & 0.2724 \\ \cline{2-5}
    & ScrVSR (Ours)              & \textbf{29.54} & \textbf{0.9609} & \textbf{0.1927} \\ \hline

    \multirow{3}{*}{QP-22} 
    & BTC\cite{btc} & 27.00 & 0.9409 & 0.2390 \\ \cline{2-5}
    & ITSRN\cite{itsrn} & 27.91 & 0.9494 & 0.2973 \\ \cline{2-5}
    & ScrVSR (Ours)              & \textbf{29.17} & \textbf{0.9568} & \textbf{0.2089} \\ \hline

    \multirow{3}{*}{QP-27} 
    & BTC\cite{btc} & 26.52 & 0.9312 & 0.2837 \\ \cline{2-5}
    & ITSRN\cite{itsrn} & 26.91 & 0.9362 & 0.3645 \\ \cline{2-5}
    & ScrVSR (Ours)              & \textbf{28.17} & \textbf{0.9465} & \textbf{0.2598} \\ \hline

    \multirow{3}{*}{QP-32} 
    & BTC\cite{btc} & 25.45 & 0.9106 & 0.4111 \\ \cline{2-5}
    & ITSRN\cite{itsrn} & 25.15 & 0.9099 & 0.4977 \\ \cline{2-5}
    & ScrVSR (Ours)              & \textbf{26.17} & \textbf{0.9239} & \textbf{0.3838} \\ \hline

    \multirow{3}{*}{QP-34} 
    & BTC\cite{btc} & 24.82 & 0.8970 & 0.4907 \\ \cline{2-5}
    & ITSRN\cite{itsrn} & 24.37 & 0.8951 & 0.5661 \\ \cline{2-5}
    & ScrVSR (Ours)              & \textbf{25.18} & \textbf{0.9096} & \textbf{0.4702} \\ \hline

    \multirow{3}{*}{QP-37} 
    & BTC\cite{btc} & \textbf{23.77} & 0.8733 & 0.6206 \\ \cline{2-5}
    & ITSRN\cite{itsrn} & 23.24 & 0.8689 & 0.6935 \\ \cline{2-5}
    & ScrVSR (Ours)              & 23.74 & \textbf{0.8835} & \textbf{0.6040} \\ \hline
    \multirow{3}{*}{Overall}
    & BTC\cite{btc} & 25.79	& 0.9163	& 0.3765  \\ \cline{2-5} 
    & ITSRN\cite{itsrn} & 25.98	& 0.9191	& 0.4486  \\ \cline{2-5} 
    & ScrVSR (Ours)   & \textbf{27.00}	& \textbf{0.9302}	& \textbf{0.3532}\\  \hline

    \end{tabular}
\end{table}

As seen in Table \ref{tab:qp_wise_comparison}, at lower compression levels (QP-17), ScrVSR achieves a PSNR of $29.54$ and SSIM of $0.9609$, representing improvements of $2.36$ and $0.0061$ over BTC\cite{btc}, and $1.23$ and $0.0061$ over ITSRN\cite{itsrn}, respectively. Notably, ScrVSR also achieves the lowest CER of $0.1927$, indicating superior text recognition performance compared to BTC ($0.2137$) and ITSRN ($0.2724$). This trend persists across medium and high compression levels (QP-22, QP-27, and QP-32), where ScrVSR maintains substantial margins over the competing methods\cite{scrvsr}. We secured \texttt{Rank-2} in the IEEE ICME Grand Challenge on Video Super-Resolution for Video Conferencing (2025), with special mention on inference time as ours \texttt{2.4s} and winner team's \texttt{23s} reported in\cite{gcreport}.

\subsection{Test-Time Adaptation for Screen Content Image Super-Resolution (SCISR-TTA)}
\textbf{Methodology}: 
SCISR-TTA adopts a region-specific two-stage adaptation strategy. In the first stage, the SR model is adapted using text-sensitive objectives that emphasize edge sharpness and character fidelity within detected text regions. In the second stage, the text-adapted model is further refined on non-text regions to suppress compression artifacts and improve visual coherence.
Textual feedback is derived from OCR and a Small Language Model (SLM). Text extracted from the super-resolved output is refined using the SLM, and the similarity between the original OCR text and its corrected version provides a self-supervised signal indicating the reliability of the reconstructed characters.

Also, for non-text regions, we use a reconstruction-based auxiliary loss guided by a cleaned reference image generated by a lightweight NAFNet\cite{nafnet} based restoration network. This module is identical to that given in the proposed methodology\cite{rwvsrqa}; it removes noise and compression artifacts while preserving the underlying image structure and is kept fixed during test-time adaptation. Let $I_{ns}$ denote the non-text region extracted from the super-resolved image using the region mask, and let $\bar{I}_{ns}$ be the corresponding cleaned output. The natural-scene loss is defined as $\mathcal{L}_{ns} = \left\| I_{ns} - \bar{I}_{ns} \right\|_2^2$.

By separately treating text and non-text regions with complementary objectives, SCISR-TTA enables stable, targeted adaptation without requiring paired high-resolution data.

\begin{figure}[ht]
    \centering
    \includegraphics[width=\linewidth]{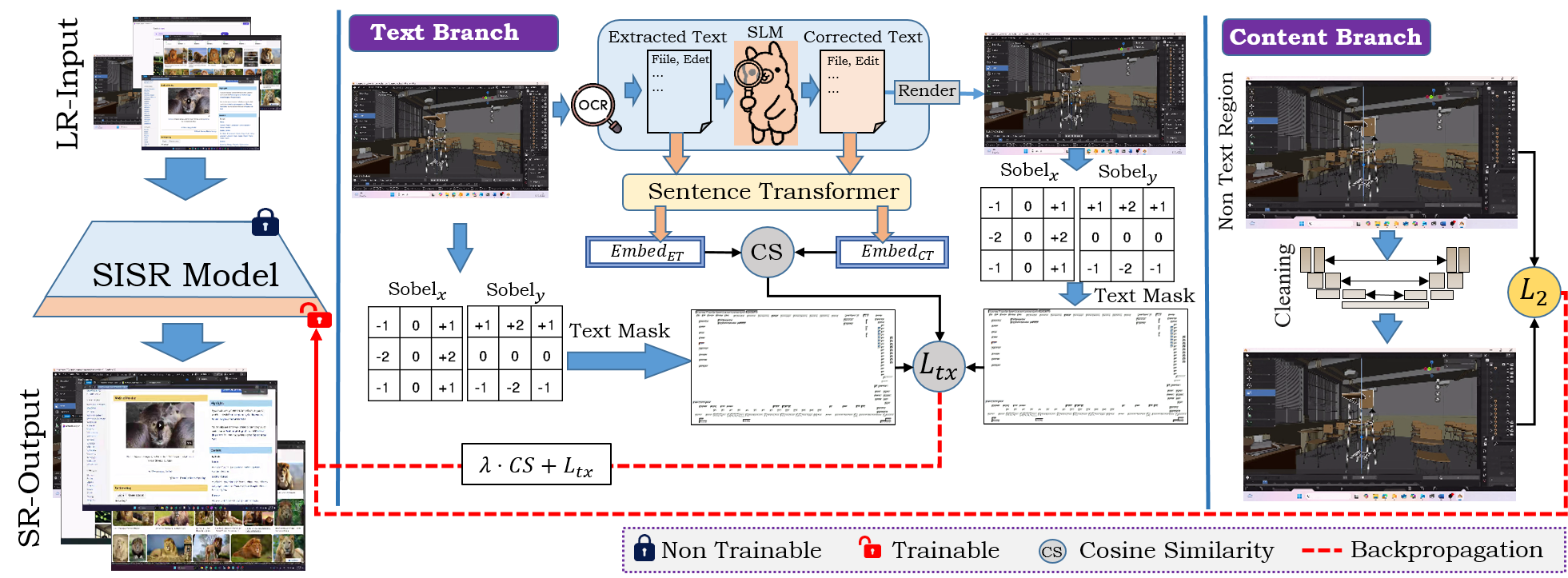}
    \caption{Overview of the proposed SCISR-TTA. The underlying model is adapted through two region-specific branches, optimized sequentially. The text adaptation branch first updates the model, followed by the content branch.}
    \label{fig:SCISR-TTA}
\end{figure}

\textbf{Results}:
Across all baselines, SCISR-TTA consistently improves both text fidelity and perceptual quality as shown in Table \ref{tab:tta_results}. For ITSRN, test-time adaptation reduces CER from $0.5762$ to $0.5621$, while simultaneously increasing DFSS\cite{dfss}, DISTS\cite{dists}, and AFINE\cite{afine}.

\begin{table}[ht]
    \caption{Performance comparison of baseline SCISR methods and their adapted versions using the proposed SCISR-TTA framework.}
    \centering
    \resizebox{\columnwidth}{!}{%
    \begin{tabular}{|l||c|c|c|c|c|}
    \hline
        \textbf{Model} & \textbf{CER} $\downarrow$ & \textbf{DFSS} $\uparrow$ & \textbf{DISTS} $\uparrow$ & \textbf{AFINE} $\uparrow$ & \textbf{EIQM} $\uparrow$ \\ \hline \hline
        
        ITSRN \cite{itsrn}    
        & 0.5762 & 46.7358 & 0.0482 & -1.3066 &  \textbf{0.1655} \\ \hline
        ITSRN\_ada            
        & \textbf{0.5621} & \textbf{47.6527} & \textbf{0.0497} & \textbf{-1.2374} & 0.1636 \\ \hline 

        BTC \cite{btc}        
        & 0.6886 & 45.5911 & 0.0565 & -1.1270 & 0.1860\\ \hline
        BTC\_ada              
        & \textbf{0.6844} & \textbf{47.5771} & \textbf{0.1150} & \textbf{-0.4674} & \textbf{0.2132} \\ \hline 

        LIIF \cite{liif}      
        & 0.7692 & 46.6831 & 0.0481 & -0.8110 &  \textbf{0.2089} \\ \hline
        LIIF\_ada             
        & \textbf{0.7648} & \textbf{47.9119} & \textbf{0.0492} & \textbf{-0.7463} & 0.2035 \\ \hline 

        ScrVSR \cite{scrvsr}  
        & 0.4001 & 46.6005 & 0.0406 & -1.7986 & 0.1065 \\ \hline
        ScrVSR\_ada           
        & \textbf{0.3778} & \textbf{47.3395} & \textbf{0.0542} & \textbf{-1.7561} & \textbf{0.1130} \\ \hline 

        LTE \cite{lte}        
        & 0.8051 & 45.4898 & 0.0485 & -0.7668 & 0.2067 \\ \hline
        LTE\_ada & \textbf{0.7669} & \textbf{49.8548} & \textbf{0.0853} & \textbf{-0.1911} &  \textbf{0.2340} \\ \hline 

    \end{tabular}
    }
    \label{tab:tta_results}
\end{table}

\noindent This indicates that SCISR-TTA enhances not only pixel-level appearance but also the recognizability of textual content, which is critical for screen images. Similar trends are observed for BTC\cite{btc} and LIIF\cite{liif}, where SCISR-TTA yields consistent reductions in CER and noticeable gains in perceptual metrics, confirming that the proposed region-aware adaptation is effective even for models not originally designed for text-sensitive reconstruction\cite{scisrtta}.

\section{Outcomes and Conclusions}
My doctoral work presented a unified framework for enhancing screen-content video super-resolution, real-world perceptual quality assessment, and quality-guided optimization. The work is accomplished in three key stages. 
First, a no-reference \textbf{Real-World VSR Quality Assessment (RW-VSR-QA)} algorithm is developed, capable of self-supervised adaptation during inference, improving perceptual alignment and robustness across six representative VSR models\cite{rwvsrqa}. 
Second, \textbf{ScrVSR}, a transformer-based super-resolution model, was developed to preserve text clarity, sharp edges, and structured graphical elements in screen-sharing videos, achieving superior PSNR, SSIM, and Character Error Rate (CER)\cite{scrvsr}. 
Finally, \textbf{SCISR-TTA}, a region-specific dual-branch test-time adaptation method, was introduced to handle text and natural-scene regions separately, improving perceptual sharpness, readability, and structural fidelity across multiple SR backbones without requiring ground-truth supervision \cite{scisrtta}.

\section{Impact Statement}
The presented work advances adaptive, no-reference video super-resolution and quality assessment, benefiting video conferencing, remote learning, and streaming platforms globally. The proposed test-time adaptation (TTA) frameworks eliminate the need for paired high-resolution data, enabling deployment across diverse real-world devices and network conditions.

\section{Issues for Discussion}

The proposed methods primarily operate on individual frames and do not explicitly model temporal dependencies. Incorporating temporal modeling to improve temporal consistency, reduce flickering, and enhance perceptual quality in real-world video super-resolution remains an important direction for future research.

\section{Acknowledgment}

This work represents the culmination of the author's doctoral research conducted at the Indian Institute of Technology Jammu. The author acknowledges \textit{Prof. Vinit Jakhetiya} (Associate Professor, IIT Jammu) for his guidance and mentorship, and the Ministry of Education (formerly MHRD), Government of India, for Fellowship.

\balance
\bibliographystyle{acm}
\bibliography{acmart}

@article{gcreport,
  title={ICME 2025 Grand Challenge on Video Super-Resolution for Video Conferencing},
  author={Naderi, Babak and Cutler, Ross and Cho, Juhee and Khongbantabam, Nabakumar and Ivkovic, Dejan},
  journal={arXiv preprint arXiv:2506.12269},
  year={2025}
}

@article{itsrn,
  title={Implicit transformer network for screen content image continuous super-resolution},
  author={Yang, Jingyu and Shen, Sheng and Yue, Huanjing and Li, Kun},
  journal={Advances in Neural Information Processing Systems},
  volume={34},
  pages={13304--13315},
  year={2021}
}

@inproceedings{btc,
  title={Compressed Screen Content Image Enhancement with B-Spline Based Distortion Estimation},
  author={Li, Junru and Li, Yue and Lin, Chaoyi and Zhang, Kai and Zhang, Li},
  booktitle={2025 Data Compression Conference (DCC)},
  pages={93--102},
  year={2025},
  organization={IEEE}
}

@inproceedings{lte,
  title={Local texture estimator for implicit representation function},
  author={Lee, Jaewon and Jin, Kyong Hwan},
  booktitle={Proceedings of the IEEE/CVF conference on computer vision and pattern recognition},
  pages={1929--1938},
  year={2022}
}

@inproceedings{liif,
  title={Learning continuous image representation with local implicit image function},
  author={Chen, Yinbo and Liu, Sifei and Wang, Xiaolong},
  booktitle={Proceedings of the IEEE/CVF conference on computer vision and pattern recognition},
  pages={8628--8638},
  year={2021}
}

@inproceedings{scrvsr,
  title={ScrVSR: Screen Sharing Video Super-Resolution},
  author={Verma, Ajeet Kumar and Tripathi, Shweta and Jakhetiya, Vinit and Subudhi, Badri N and Jaiswal, Sunil},
  booktitle={2025 IEEE International Conference on Multimedia and Expo Workshops (ICMEW)},
  pages={1--6},
  year={2025},
  organization={IEEE}
}

@inproceedings{real_basic_2022,
  title = {Investigating Tradeoffs in Real-World Video Super-Resolution},
  author = {K. Chan and S. Zhou and X. Xu and C. C. Loy},
  booktitle = {IEEE Conference on Computer Vision and Pattern Recognition},
  year = {2022}
}

@article{tecoGAN, 
  title={Learning Temporal Coherence via Self-Supervision for GAN-based Video Generation (TecoGAN)}, 
  author={M. Chu and Y. Xie and M. Jonas and L. Laura and N. Thuerey}, 
  journal={ACM Transactions on Graphics (TOG)},
  volume={39},
  number={4},
  year={2020},
  publisher={ACM} 
}

@InProceedings{realesrgan,
  title     = {Real-ESRGAN: Training Real-World Blind Super-Resolution with Pure Synthetic Data},
  author    = {X. Wang and L. Xie and C. Dong and Y. Shan},
  booktitle = {International Conference on Computer Vision Workshops (ICCVW)},
  year      = {2021}
}

@inproceedings{basicvsrpp,
  title = {{BasicVSR++}: Improving video super-resolution with enhanced propagation and alignment},
  author = {K. Chan and S. Zhou and X. Xu and C. Loy},
  booktitle = {IEEE Conference on Computer Vision and Pattern Recognition},
  year = {2022}
}

@inproceedings{fast_vqa,
  title={Fast-vqa: Efficient end-to-end video quality assessment with fragment sampling},
  author={Wu, Haoning and Chen, Chaofeng and Hou, Jingwen and Liao, Liang and Wang, Annan and Sun, Wenxiu and Yan, Qiong and Lin, Weisi},
  booktitle={Computer Vision--ECCV 2022},
  pages={538--554},
  year={2022},
  organization={Springer}
}

@article{2022faster_vqa,
  title={Neighbourhood representative sampling for efficient end-to-end video quality assessment},
  author={Wu, Haoning and Chen, Chaofeng and Liao, Liang and Hou, Jingwen and Sun, Wenxiu and Yan, Qiong and Gu, Jinwei and Lin, Weisi},
  journal={IEEE Transactions on Pattern Analysis and Machine Intelligence},
  year={2023},
  publisher={IEEE}
}

@inproceedings{dover,
  title={Exploring video quality assessment on user generated contents from aesthetic and technical perspectives},
  author={Wu, Haoning and Zhang, Erli and Liao, Liang and Chen, Chaofeng and Hou, Jingwen and Wang, Annan and Sun, Wenxiu and Yan, Qiong and Lin, Weisi},
  booktitle={Proceedings of the IEEE/CVF International Conference on Computer Vision},
  pages={20144--20154},
  year={2023}
}

@inproceedings{sama,
  title={Scaling and masking: A new paradigm of data sampling for image and video quality assessment},
  author={Liu, Yongxu and Quan, Yinghui and Xiao, Guoyao and Li, Aobo and Wu, Jinjian},
  booktitle={Proceedings of the AAAI Conference on Artificial Intelligence},
  volume={38},
  pages={3792--3801},
  year={2024}
}

@inproceedings{cover,
  title={COVER: A comprehensive video quality evaluator},
  author={He, Chenlong and Zheng, Qi and Zhu, Ruoxi and Zeng, Xiaoyang and Fan, Yibo and Tu, Zhengzhong},
  booktitle={Proceedings of the IEEE/CVF Conference on Computer Vision and Pattern Recognition},
  pages={5799--5809},
  year={2024}
}

@inproceedings{eiqm,
  title={Enhancing image quality prediction with self-supervised visual masking},
  author={{\c{C}}o{\u{g}}alan, U{\u{g}}ur and Bemana, Mojtaba and Seidel, Hans-Peter and Myszkowski, Karol},
  booktitle={Computer Graphics Forum},
  volume={43},
  pages={e15051},
  year={2024},
  organization={Wiley Online Library}
}

@inproceedings{afine,
  title={Toward generalized image quality assessment: Relaxing the perfect reference quality assumption},
  author={Chen, Du and Wu, Tianhe and Ma, Kede and Zhang, Lei},
  booktitle={Proceedings of the Computer Vision and Pattern Recognition Conference},
  pages={12742--12752},
  year={2025}
}

@article{dists,
  title={Image quality assessment: Unifying structure and texture similarity},
  author={Ding, Keyan and Ma, Kede and Wang, Shiqi and Simoncelli, Eero P},
  journal={IEEE Transactions on Pattern Analysis and Machine Intelligence},
  volume={44},
  number={5},
  pages={2567--2581},
  year={2020},
  publisher={IEEE}
}

@article{rwvsrqa,
  title={Real-World Video Quality Assessment via Test-Time Adaptation and its application in Real-World Video Super-Resolution},
  author={Verma, Ajeet Kumar and Mishra, Ambuj and Jakhetiya, Vinit and Subudhi, Badri Narayan and Jaiswal, Sunil},
  journal={IEEE Transactions on Artificial Intelligence},
  year={2025},
  publisher={IEEE}
}

@article{dfss,
  title={Deep feature statistics mapping for generalized screen content image quality assessment},
  author={Chen, B and Zhu, H and Zhu, L and Wang, S and Kwong, S},
  journal={IEEE Transactions on Image Processing},
  volume={33},
  pages={3227--3241},
  year={2024},
  publisher={IEEE}
}

@inproceedings{arniqa,
  title={Arniqa: Learning distortion manifold for image quality assessment},
  author={Agnolucci, L and Galteri, L and Bertini, M and Del B, A},
  booktitle={Proceedings of the IEEE/CVF Winter Conference on Applications of Computer Vision},
  pages={189--198},
  year={2024}
}

@inproceedings{iart,
  title={Enhancing Video Super-Resolution via Implicit Resampling-based Alignment},
  author={Xu, Kai and Yu, Ziwei and Wang, Xin and Mi, Michael Bi and Yao, Angela},
  booktitle={Proceedings of the IEEE/CVF Conference on Computer Vision and Pattern Recognition},
  pages={2546--2555},
  year={2024}
}

@inproceedings{nafnet,
  title={Simple baselines for image restoration},
  author={Chen, Liangyu and Chu, Xiaojie and Zhang, Xiangyu and Sun, Jian},
  booktitle={European conference on computer vision},
  pages={17--33},
  year={2022},
  organization={Springer}
}

@inproceedings{srwd,
  title={Expanding synthetic real-world degradations for blind video super resolution},
  author={Jeelani, Mehran and Cheema, Noshaba and Illgner-Fehns, Klaus and Slusallek, Philipp and Jaiswal, Sunil and others},
  booktitle={Proceedings of the IEEE/CVF conference on computer vision and pattern recognition},
  pages={1199--1208},
  year={2023}
}

@inproceedings{edsr,
  title={Enhanced deep residual networks for single image super-resolution},
  author={Lim, Bee and Son, Sanghyun and Kim, Heewon and Nah, Seungjun and Mu Lee, Kyoung},
  booktitle={Proceedings of the IEEE conference on computer vision and pattern recognition workshops},
  pages={136--144},
  year={2017}
}

@inproceedings{hat,
  title={Activating more pixels in image super-resolution transformer},
  author={Chen, Xiangyu and Wang, Xintao and Zhou, Jiantao and Qiao, Yu and Dong, Chao},
  booktitle={Proceedings of the IEEE/CVF conference on computer vision and pattern recognition},
  pages={22367--22377},
  year={2023}
}

@inproceedings{scisrtta,
  title={SCISR-TTA: Test Time Adaptation for Screen Content Image Super Resolution},
  author={Verma, Ajeet Kumar and Jakhetiya, Vinit and Subudhi, Badri N},
  booktitle={2026 IEEE International Conference on Multimedia and Expo Workshops (ICMEW)},
  pages={1--6},
  year={2026},
  organization={IEEE}
}

\end{document}